\pdfoutput=1

\documentclass[11pt]{article}

\usepackage{acl}

\usepackage{times}
\usepackage{latexsym}
\usepackage{amsmath}

\usepackage[T1]{fontenc}

\usepackage[utf8]{inputenc}
\usepackage{multirow}
\usepackage{multicol}
\usepackage{booktabs}
\usepackage{subfigure}
\usepackage{multirow}
\usepackage{ragged2e,array}
\usepackage{microtype}
\usepackage{bbding}
\usepackage{graphicx}
\usepackage{enumitem}
\usepackage{makecell}

\title{MOVER: Mask, Over-generate and Rank for Hyperbole Generation}

\author {
    Yunxiang Zhang,
    Xiaojun Wan \\
    Wangxuan Institute of Computer Technology, Peking University\\
    The MOE Key Laboratory of Computational Linguistics, Peking University\\
    \texttt{\{yx.zhang,wanxiaojun\}@pku.edu.cn}
}

\begin{document}
\maketitle
\begin{abstract}
Despite being a common figure of speech, hyperbole is under-researched in Figurative Language Processing. In this paper, we tackle the challenging task of hyperbole generation to transfer a literal sentence into its hyperbolic paraphrase. To address the lack of available hyperbolic sentences, we construct HYPO-XL, the first large-scale English hyperbole corpus containing 17,862 hyperbolic sentences in a non-trivial way. Based on our corpus, we propose an unsupervised method for hyperbole generation that does not require parallel literal-hyperbole pairs. During training, we fine-tune BART \cite{lewis-etal-2020-bart} to infill masked hyperbolic spans of sentences from HYPO-XL. During inference, we mask part of an input literal sentence and over-generate multiple possible hyperbolic versions. Then a BERT-based ranker selects the best candidate by hyperbolicity and paraphrase quality. Automatic and human evaluation results show that our model is effective at generating hyperbolic paraphrase sentences and outperforms several baseline systems.
\end{abstract}

\section{Introduction}

Hyperbole is a figure of speech that deliberately exaggerates a claim or statement to show emphasis or express emotions. If a referent has a feature X, a hyperbole exceeds the credible limits of fact in the given context and presents it as having more of that X than warranted by reality \citep{claridge2010hyperbole}. Take the following example, ``\textit{I won’t wait for you: it took you centuries to get dressed.}'' It over-blows the time for someone to get dressed with a single word ``\textit{centuries}'' and thus creates a heightened effect. From a syntactic point of view, \citet{claridge2010hyperbole} classifies hyperbole into word-level, phrase-level and clause-level types, and conclude that the former two types are more common in English. Although hyperbole is considered as the second most frequent figurative device \citep{KREUZ1993151}, it has received less empirical attention in the NLP community. Recently \citet{tian-etal-2021-hypogen-hyperbole} addressed the generation of \textit{clause-level} hyperbole.
In this paper, we instead focus on \textit{word-level} and \textit{phrase-level} hyperbole, which can be unified as span-level hyperbole. 



To tackle the hyperbole generation problem we need to address three main challenges:
\begin{itemize}[itemsep=2pt,topsep=2pt,parsep=2pt]
    \item The lack of training data that either consists of large-scale hyperbolic sentences or literal-hyperbole pairs, which are necessary to train an unsupervised or supervised model.
    \item The tendency of generative language models to produce literal text rather than hyperboles.
    \item Trade-off between content preservation and hyperbolic effect of the generated sentences.
\end{itemize}

In order to address the above challenges, we propose \textbf{MOVER} (\textbf{M}ask, \textbf{OVE}r-generate and \textbf{R}ank), an unsupervised approach to generating hyperbolic paraphrase from literal input. Our approach does not require parallel data for training, thus alleviating the issue of scarce data. Still, we need a non-parallel corpus containing as much hyperbolic sentences as possible. To this end, we first build a large-scale English hyperbole corpus HYPO-XL in a weakly supervised way. 

Based on the intuition that the hyperbolic effect of a sentence is realized by a single word or phrase within it, we introduce a sub-task of hyperbolic span extraction. We identify several possible n-grams of a hyperbolic sentence that can cause the hyperbolic bent with syntactic and semantic features. We apply this masking approach to sentences in HYPO-XL and teach a pretrained seq2seq transformer, BART \citep{lewis-etal-2020-bart}, to infill the words in missing hyperbolic spans. This increases the probability of generating hyperbolic texts instead of literal ones. During inference, given a single literal sentence, our system provides multiple masked versions for inputs to BART and generates potential hyperbolic sentences accordingly. To select the best one for output, we leverage a BERT-based ranker to achieve a satisfying trade-off between hyperbolicity and paraphrase quality.

Our contributions are three-fold:
\begin{itemize}[itemsep=2pt,topsep=2pt,parsep=2pt]
\item We construct the first large-scale hyperbole corpus HYPO-XL in a non-trivial way. The corpus is publicly available,\footnote{Code and data are available at \url{https://github.com/yunx-z/MOVER}.} contributing to the Figurative Language Processing (FLP) community by facilitating the development of computational study of hyperbole.
\item  We propose an unsupervised approach for hyperbole generation that falls into the ``overgenerate-and-rank'' paradigm \citep{heilman2009question}.
\item We benchmark our system against several baselines and we compare their performances by pair-wise manual evaluations to demonstrate the effectiveness of our approach.
\end{itemize}

\section{HYPO-XL: Hyperbole Corpus Collection}\label{sec:hypo-xl-crea}

The availability of large-scale corpora can facilitate the development of figurative language generation with pretrained models, as is shown by \citet{chakrabarty-etal-2020-generating} on simile generation and \citet{chakrabarty2021mermaid} on metaphor generation. However, datasets for hyperbole are scarce.  \citet{troiano-etal-2018-computational} built an English corpus HYPO containing 709 triplets $[hypo, para, non\_hypo]$, where $hypo$ refers to a hyperbolic sentence, $para$ denotes the literal paraphrase of $hypo$ and $non\_hypo$ means a non-hyperbolic sentence that contains the same hyperbolic word or phrase as $hypo$ but with a literal connotation. 
The size of this dataset is too small to train a deep learning model for hyperbole detection and generation. To tackle the lack of hyperbole data, we propose to enlarge the hyperbolic sentences of HYPO in a weakly supervised way and build a large-scale English corpus of 17,862 hyperbolic sentences, namely HYPO-XL. We would like to point out that this is a \textit{non-parallel} corpus containing only hyperbolic sentences without their paraphrase counterparts, because our hyperbole generation approach (Section \ref{hypo-gen}) does not require parallel training data. 

The creation of HYPO-XL consists of two steps:
\begin{enumerate}[itemsep=2pt,topsep=2pt,parsep=2pt]
\item We first train a BERT-based binary classifier on HYPO and retrieve possible hyperbolic sentences from an online corpus. \item We manually label a subset of the retrieved sentences, denoted HYPO-L, and retrain our hyperbole detection model to identify hyperbolic sentences from the same retrieval corpus with higher confidence.
\end{enumerate}

\subsection{Automatic Hyperbole Detection}\label{hypo-detect}
Hyperbole detection is a supervised binary classification problem where we predict whether a sentence is hyperbolic or not \citep{kong-etal-2020-identifying}. We fine-tune a BERT-base model \citep{devlin-etal-2019-bert} on the hyperbole detection dataset HYPO \citep{troiano-etal-2018-computational}. 
In experiment, we randomly split the data into 567 (80\%) hyperbolic sentences, with their literal counterparts ($para$ and $non\_hypo$) as negative samples, in training set and 71 (10\%) in development set and 71 (10\%) in test set. Our model achieves an accuracy of 80\%
on the test set, which is much better than the highest reported accuracy (72\%) of traditional algorithms in \citet{troiano-etal-2018-computational}.

Once we obtain this BERT-based hyperbole detection model, the next step is to retrieve hyperbolic sentences from a corpus. Following \citet{chakrabarty-etal-2020-r}, we use Sentencedict.com,\footnote{\url{https://sentencedict.com/}} an online sentence dictionary as the retrieval corpus.
We remove duplicate and incomplete sentences (without initial capital) in the corpus, resulting in a collection of 767,531 sentences. Then we identify 93,297 (12.2 \%) sentences predicted positive by our model as pseudo-hyperbolic.

\subsection{HYPO-L: Human Annotation of Pseudo-labeled Data}\label{human-anno}

Due to the small size of training set, pseudo-labeled data tend to have lower confidence score (i.e., the prediction probability). To improve the precision of our model,\footnote{Given the massive hyperboles in the ``wild'' (i.e., the retrieval corpus) we do not pursue recalling
more hyperboles at the risk of hurting precision \cite{zhang2021writing}. 
} we further fine-tune it with our human-annotated data, namely HYPO-L. We randomly sample 5,000 examples from the 93,297 positive predictions and invite students with proficiency in English to label them as hyperbolic or not. For each sentence, two annotators provide their judgements. We only keep items with unanimous judgments (i.e. both of the two annotators mark the sentence as hyperbolic or non-hyperbolic) to ensure the reliability of annotated data. In this way, 3,226 (64.5\%) out of 5,000 annotations are left in HYPO-L. This percentage of unanimous judgments (i.e., raw agreement, RA) is comparable to  58.5\% in the creation of HYPO \cite{troiano-etal-2018-computational}. To be specific, HYPO-L consists of 1,007 (31.2\%) hyperbolic sentences (positive samples) and 2,219 (68.8\%) literal ones (negative samples). 

We continue to train the previous HYPO-fine-tuned BERT on HYPO-L and the test accuracy is 80\%,\footnote{We separate 10\% data of HYPO-L for development and another 10\% for testing.} which we consider as an acceptable metric for hyperbole detection.  Finally we apply the BERT-based detection model to the retrieval corpus again and retain sentences whose prediction probabilities for positive class exceed a certain threshold.\footnote{Based on manual inspection of predicted results, we set the threshold as 0.8 to trade-off between precision and recall.}
This results in HYPO-XL, a large-scale corpus of 17,862 (2.3\%) hyperbolic sentences.
We provide a brief comparison of HYPO, HYPO-L and HYPO-XL in Table \ref{hypo-comp} to further clarify the data collection process.

\begin{table}
\setlength\tabcolsep{4pt}
\centering
\begin{tabular}{lcccc}
\toprule
Dataset & \# Hypo. & \# Non. & \# Para. & \# Total \\
\hline
HYPO  & 709 & 698 & 709 & 2,116 \\ 
HYPO-L  & 1,007 & 2,219  & - & 3,226 \\
HYPO-XL  & 17,862 & - & - & 17,862 \\
\bottomrule
\end{tabular}
\caption{Comparision of different hyperbole datasets (corpora) in terms of hyperbolic (Hypo.), non-hyperbolic (Non.) and paraphrase (Para.) sentences.
}\label{hypo-comp}
\end{table}

\begin{table}
\centering
\begin{tabular}{lcc}\toprule
Measurement &Value \\\cmidrule{1-2}
\% Non-hypo & 6\% \\\cmidrule{1-2}
\# Avg hypo span tokens &2.23 \\
\% Long hypo spans ($>$ 1 token) &37\% \\
\# Distinct hypo spans &85 \\
\# Distinct POS-ngrams of hypo spans &39 \\
\bottomrule
\end{tabular}
\caption{Statistics of 100 random samples from HYPO-XL, of which 6 are actually non-hyperboles (``Non-hypo''). The statistics of hyperbolic text spans (``hypo span'') are calculated for the rest 94 real hyperboles.}\label{tab:hypo-xl-stat}
\end{table}

\subsection{Corpus Analysis}\label{sec:corpus-analysis}

Since HYPO-XL is built in a weakly supervised way with only a few human labeled data samples, we conduct a quality analysis to investigate how many sentences in the corpus are \textit{actually} hyperbolic. We randomly sample 100 instances from HYPO-XL and manually label them as hyperbole or non-hyperbole. Only six sentences are not hyperbole. This precision of 94\% is on par with 92\% on another figurative language corpus of simile \cite{zhang2021writing}. Actually we can tolerate a bit noise in the corpus since the primary goal of HYPO-XL is to facilitate hyperbole \textit{generation} instead of \textit{detection}, and a small proportion of non-hyperbole sentences as input will not harm our proposed method.\footnote{We further explain the reason in Section \ref{hypo-bart}} 
Table \ref{tab:hypo-xl-stat} shows the statistics of hyperbolic text  spans (defined in Section \ref{hypo-span}) for the rest 94 real hyperboles. We also provide additional analyses 
in Appendix \ref{sec:dataset-stat}.

\section{Hyperbole Generation}\label{hypo-gen}

  \begin{figure*}
    \includegraphics[width=\linewidth]{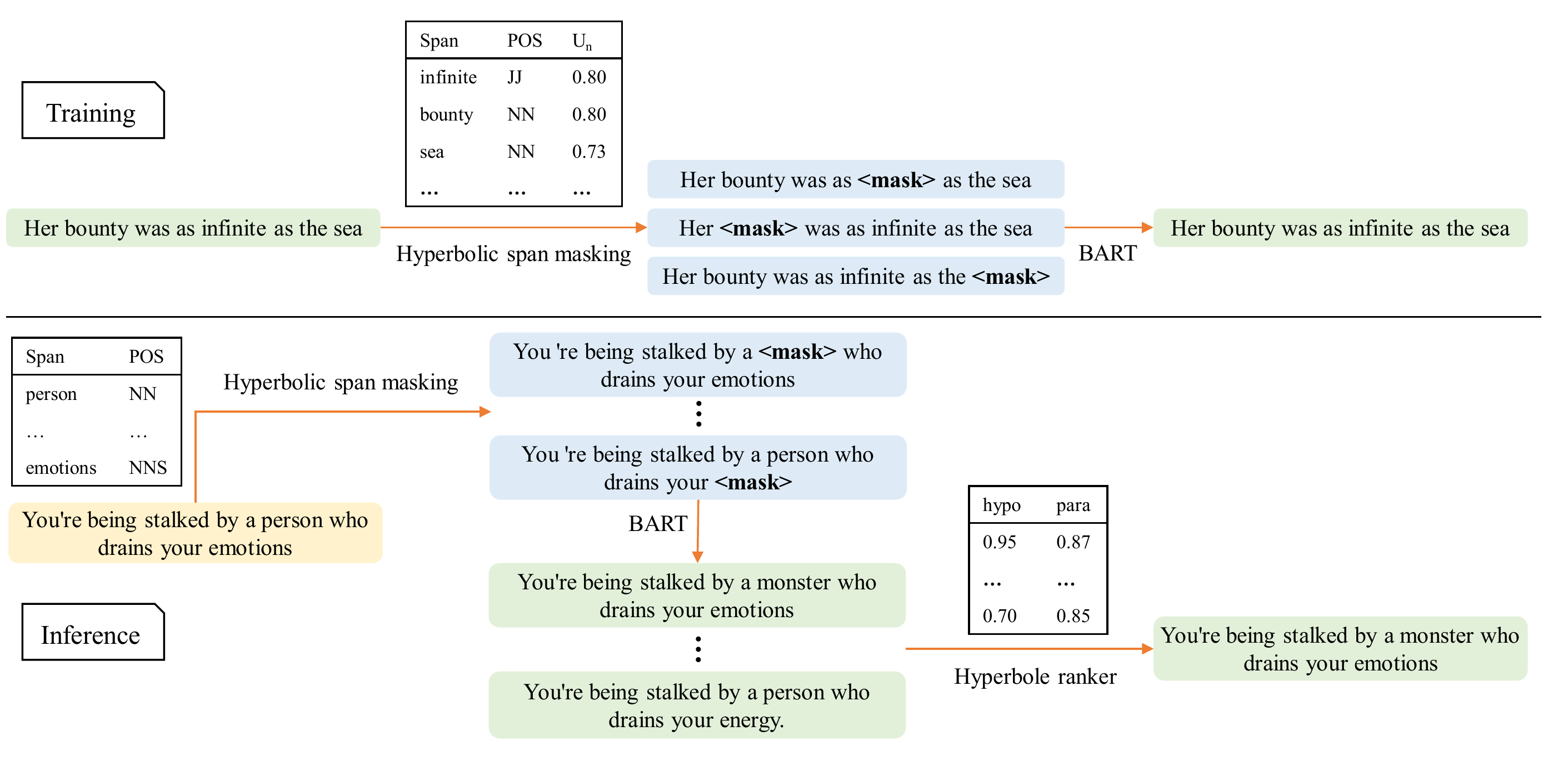}
    \caption{Overview of our approach to unsupervised hyperbole generation. Literal sentences are in yellow boxes, masked sentences are in blue boxes and hyperbolic sentences are in green boxes.}\label{fig:mover}
  \end{figure*}

We propose an unsupervised approach to generate hyperbolic paraphrase from a literal sentence with BART \citep{lewis-etal-2020-bart} such that we do not require parallel literal-hyperbole pairs.\footnote{We note that training our model still relies on instances from a specialized hyperbole corpus.} An overview of our hyperbole generation pipeline is
shown in Figure \ref{fig:mover}. It consists of two steps during training:
\begin{enumerate}[itemsep=2pt,topsep=3pt,parsep=2pt]
\item \textbf{Mask.} Given a hyperbolic sentence from HYPO-XL, we identify multiple text spans that can possibly produce the hyperbolic meaning of a sentence, based on two features (POS n-gram and unexpectedness score). For each identified text span, we replace it with the \texttt{<mask>} token to remove hyperbolic attribute of the input. $N$ text spans will result in $N$ masked inputs, respectively. 
\item \textbf{Infill.}  We fine-tune BART to fill the masked spans of input sentences. The model learns to generate hyperbolic words or phrases that are pertinent to the context.
\end{enumerate}

During inference, there are three steps:
\begin{enumerate}[itemsep=2pt,topsep=3pt,parsep=2pt]
\item \textbf{Mask.} Given a literal sentence, we apply POS-ngram-only masking to produce multiple input sentences. 
\item \textbf{Over-generate.} BART generates one sentence from a masked input, resulting in multiple candidates. 
\item \textbf{Rank.} Candidates are ranked by their hyperbolicity and relevance to the source literal sentence. The one with highest score is selected as the final output.
\end{enumerate}

We dub our hyperbole generation system \textbf{MOVER} (\textbf{M}ask, \textbf{OVE}r-generate and \textbf{R}ank). We apply masking technique to map both the hyperbolic (training input) and literal (test input) sentences into a same ``space'' where the masked sentence can be transformed into hyperbole by BART. It falls into the ``overgenerate-and-rank'' paradigm \citep{heilman2009question} since
many candidates are available after the generation step. The remainder of this section details the three main modules: hyperbolic span masking (Section \ref{hypo-span}), BART-based span infilling (Section \ref{hypo-bart}) and the hyperbole ranker (Section \ref{hypo-rank}).

\subsection{Mask: Hyperbolic Span Masking}\label{hypo-span}

 We make a simple observation that the hyperbolic effect of a sentence is commonly localized to a single word or a phrase, which is also supported by a corpus-based linguistic study on hyperbole \citep{claridge2010hyperbole}. For example, the word \textit{marathon} in \textit{``My evening jog with Bill turned into a \textbf{marathon}''} overstates the jogging distance and causes the sentence to be hyperbolic. This inspires us to leverage the ``delete-and-generate'' strategy \citep{li-etal-2018-delete} for hyperbole generation. 
 Concretely, a literal sentence can be transformed into its hyperbolic counterpart via hyperbolic span extraction and replacement.
 We propose to extract hyperbolic spans based on POS n-gram  (syntactic) and unexpectedness (semantic) features. 
 
\paragraph{POS N-gram} 
We extract POS n-gram patterns of hyperbole from the training set of HYPO dataset\footnote{The hyperbolic spans are not explicitly provided in the HYPO dataset, so we take the maximum word overlap between $hypo$ and $non\_hypo$ (Section \ref{sec:hypo-xl-crea}) as the hyperbolic spans.} and obtain 262 distinct POS n-grams. As a motivating example, the following three hyperbolic spans, \textit{``faster than light'', ``sweeter than honey'', ``whiter than snow''}, share the same POS n-gram of ``JJR+IN+NN''.

\begin{figure}
\includegraphics[width=0.9\linewidth]{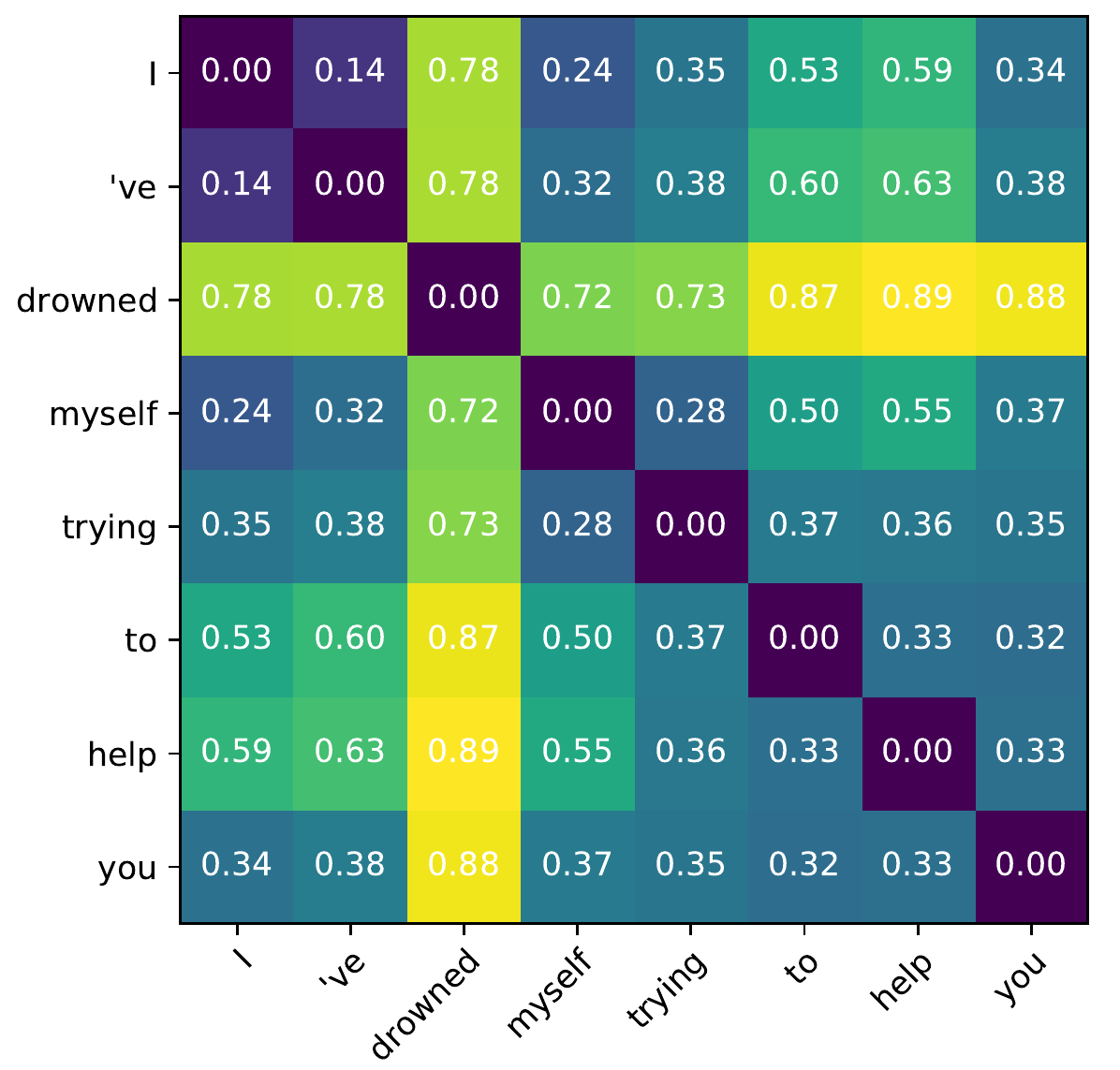}
\caption{A visualization of the cosine distance matrix of the  hyperbolic sentence ``\textit{I've drowned myself trying to help you}''. 
}\label{fig:unexpect}
\end{figure}

\paragraph{Unexpectedness} Hyperbolic spans are less coherent with the literal contexts and thus their vector representations are distant from the context vectors.  \citet{troiano-etal-2018-computational} have verified this intuition with the unexpectedness metric. They define the unexpectedness score $U_s$ of a sentence $s$ with the token sequence $\{x_0, x_1, ..., x_N\}$ as the average cosine distance among all of its word pairs.
 \begin{equation}
     U_s = \mathop{\text{average}}_{i,j \in [0,N], i \neq j}(\text{cosine\_distance}(v_i, v_j))
 \end{equation}
\noindent where $v_i$ denotes the word embedding vector of token $x_i$.  
Similarly, we define the unexpectedness score $U_n$ of an n-gram $\{x_k, x_{k+1}, ..., x_{k+n-1}\}$ in a sentence $s$ as the average cosine distance among word pairs that consist of one word inside the n-gram and the other outside.
  \begin{equation}
     U_n = \mathop{\text{average}}_{i \in [k,k+n-1] \atop j \in [0,k-1] \cup [k+n, N]}(\text{cosine\_distance}(v_i, v_j))
 \end{equation}
Text spans with higher unexpectedness scores tend to be hyperbolic. Figure \ref{fig:unexpect} illustrates the cosine distance of the word pairs in the sentence ``\textit{I've drowned myself trying to help you}''. The words in the span ``\textit{drowned myself}'' are distant from other words in terms of word embedding similarity.

For the masking step during training, we extract all text spans in the original input hyperbolic sentences that match one of the hyperbolic POS n-grams. Then we rank them by their unexpectedness scores and choose top-3 items as the masked spans.\footnote{This means that at least 2/3 of the identified spans should not be hyperbolic, but this will not harm the training of our hyperbole generation model, which is explained in Section \ref{hypo-bart}} For the masking step during inference, we simply mask all the spans that match hyperbolic POS n-grams, since the span unexpectedness score is not applicable to a literal input. We evaluate the accuracy of our hyperbolic span masking approach on the development set of the HYPO dataset. The proportion of exact match (EM) \citep{rajpurkar-etal-2016-squad} between our top-3 masked spans with the human-labeled spans is 86\%, which shows that our simple method based on the above-mentioned hand-crafted features is effective for the task of hyperbolic span extraction.

\subsection{Over-generate: Hyperbolic Text Infilling with  BART}\label{hypo-bart}
In order to generate hyperbolic and coherent text from the masked span, we leverage the text span infilling ability of BART \citep{lewis-etal-2020-bart}, a pretrained sequence2sequence model with a denoising autoencoder and an autoregressive autodecoder. During its pretraining, it learns to reconstruct the corrupted noised text. One of the noising transformations is random span masking, which teaches BART to predict the multiple tokens missing from a span. During our training process, we fine-tune BART by treating the masked hyperbolic sentence as the encoder source and the original one as the decoder target. This can change the probability distribution when decoding tokens and increase the chance of generating a hyperbolic, rather than literal, text span conditioned on the context. During inference, BART fills the masked span of a literal sentence with possible hyperbolic words. 

Note that if the masked span of an input sentence is actually not hyperbolic, then fine-tuning on this example will just enhance the reconstruction ability of BART, which will not exert negative effects on hyperbole generation. This can give rise to our tolerance for non-hyperbolic sentences in the training corpus (Section \ref{sec:corpus-analysis}) and non-hyperbolic masked span (Section \ref{hypo-span}). 

\subsection{Rank: Hyperbole Ranker}\label{hypo-rank}
Recall that for each literal input during inference, we apply POS-ngram-based masking, produce different masked versions of the sentence, and generate multiple output candidates. Obviously, not all masking spans are suitable for infilling hyperbolic words due to the noise of masking. To select the best candidate for final output, we introduce a hyperbole ranker which sorts candidate sentences by their degree of hyperbolicity and relevance to the source inputs.  For hyperbolicity evaluation, we leverage the BERT-based hyperbole detection model fine-tuned on HYPO and HYPO-L (Section \ref{human-anno}) to assign a hyperbole score (i.e., prediction probability) for each candidate. For the evaluation of content preservation, we train a pairwise model to predict whether hyperbolic sentence A is a paraphrase of a literal sentence B. To this end, we use the distilled RoBERTa-base model checkpoint\footnote{\url{https://huggingface.co/sentence-transformers/paraphrase-distilroberta-base-v1}} pretrained on large scale paraphrase data provided by Sentence-Transformer \citep{reimers-2019-sentence-bert}. It calculates the cosine similarity between the literal input and the candidate as the paraphrase score. We fine-tune the checkpoint on the training set of the HYPO dataset, where we treat the pairs of $hypo$ and $para$ as positive examples, and pairs of $hypo$ and $non\_hypo$ as negative examples (Section \ref{sec:hypo-xl-crea}). The accuracy on test set is 93\%.

Now that we obtain the hyperbole score $\text{hypo}(c)$ and the paraphrase score $\text{para}(c)$ for candidate $c$, we propose an intuitive scoring function $\text{score}(\cdot)$ as below:
\begin{equation}
    \text{score}(s) = 
    \begin{cases}
    \text{hypo}(s) & {\text{para}(s) \in (\gamma,1-\epsilon)} \\
    0 & \text{else}
    \end{cases}
\end{equation}
\noindent Here we filter out a candidate if its paraphrase score is lower than a specific threshold $\gamma$ or it is almost the same as the original input (i.e., the paraphrase score is extremely close to 1). For diversity purposes, we do not allow our system to simply copy the literal input as its output. We then rank the remaining candidates according to their hyperbole score and select the best one as the final output.\footnote{If all candidates are filtered out by their paraphrase scores (i.e. they all have the zero final scores), we will select the one with the highest hyperbole score among all candidates.}

\section{Experiments}\label{sec:experiments}
There are no existing models applied to the task of word- or phrase-level hyperbole generation. To compare the quality of the generated hyperboles, we benchmark our MOVER system against three baseline systems adapted from related tasks.

\subsection{Baseline Systems}

\paragraph{Retrieve (R1)} Following \citet{nikolov2020rapformer}, we implement a simple information retrieval baseline, which retrieves the closest hyperbolic sentence as the output (i.e., the highest cosine similarity) from HYPO-XL, using the hyperbole paraphrase detection model $\text{para}(\cdot)$ in Section \ref{hypo-rank}. 
The outputs of this baseline system should be hyperbolic yet have limited relevance to the input.

\paragraph{Retrieve, Replace and Rank (R3)} We first retrieve the top-5 most similar sentences from HYPO-XL like the R1 baseline. Then we apply hyperbolic span extraction in Section \ref{hypo-span}  to find 3 text spans for each retrieved sentence. We replace the text spans in a literal input sentence with retrieved hyperbolic spans if two spans share the same POS n-gram. Since this replacement method may result in multiple modified sentences, we select the best one with the hyperbole ranker in Section \ref{hypo-rank}. If there are no matched text spans, we fall back to R1 baseline and return the most similar retrieved sentence verbatim. In fact, this baseline substitutes the BART generation model in MOVER system with a simpler retrieval approach, which can demonstrate the hyperbole generation ability of BART.

\paragraph{BART} Inspired by \citet{chakrabarty-etal-2020-generating}, we replace the text infilling model in Section \ref{hypo-bart} with a non-fintuned off-the-shelf BART,\footnote{We also tried to fine-tune BART on the 567 literal-hyperbole pairs from the training set of HYPO dataset in an end-to-end supervised fashion, but the model just copy the input for all instances (same as COPY in Table \ref{auto-metric}) and is unable to generate meaningful output due to small amount of training data. Besides, we test the performance of a BART-based paraphrase generation model, which is BART finetuned on QQP \cite{wang-etal-2018-glue} and PAWS \cite{zhang2019paws} datasets. We still find that 50\% of the outputs from the paraphrase model just copy the input. Therefore, we do not consider these two BART-based systems hereafter.} because BART has already been pretrained to predict tokens from a masked span.

\subsection{Implementation Details}\label{sec:implement}
We use 16,075 (90\%) samples in HYPO-XL for training our MOVER system and the rest 1,787 sentences for validation. For POS Tagging in Section \ref{hypo-span}, we use Stanford CoreNLP \citep{manning-etal-2014-stanford}. For the word embedding, we use 840B 300-dimension version of GloVe vectors \citep{pennington-etal-2014-glove}. For BART in Section \ref{hypo-bart}, we use the BART-base checkpoint instead of BART-large due to limited computing resources and leverage the implementation by Huggingface \citep{wolf-etal-2020-transformers}. We fine-tune pretrained BART for 16 epochs. For the parameters of the hyperbole ranker in Section \ref{hypo-rank}, we set $\gamma = 0.8$ and $\epsilon = 0.001$ by manual inspection of the ranking results on the development set of the HYPO dataset.

\subsection{Evaluation Criteria}
\paragraph{Automatic Evaluation} BLEU \citep{papineni2002bleu} reflects the lexical overlap between the generated and the ground-truth text. BERTScore \citep{zhang2019bertscore} computes the similarity using contextual embeddings. These are common metrics for text generation. We use the 71 literal sentences ($para$) in the test set of HYPO dataset as test inputs and their corresponding hyperbolic sentences ($hypo$) as gold references. We report the BLEU and BERTScore metrics for generated sentences compared against human written hyperboles.

\paragraph{Human Evaluation} Automated metrics are not reliable on their own to evaluate methods to generate figurative language \citep{novikova-etal-2017-need} so we also conduct pair-wise comparisons manually \cite{shao-etal-2019-long}. We evaluate the generation results from the 71 testing literal sentences. Each pair of texts (ours vs. a baseline / human reference) is given preference (win, lose or tie) by five people with proficiency in English. We use a set of four criteria adapted from \citet{chakrabarty2021mermaid} to evaluate the generated outputs: 1) \textbf{Fluency (Flu.)}: Which sentence is more fluent and grammatical? 2) \textbf{Hyperbolicity (Hypo.)}: Which sentence is more hyperbolic? 3) \textbf{Creativity (Crea.)}: Which sentence is more creative? 4) \textbf{Relevance (Rel.)}: Which sentence is more relevant to the input literal sentence?

\begin{table}
\centering
\begin{tabular}{lcc}
\toprule
System & BLEU & BERTScore  \\
\cmidrule{1-3}
R1 & 2.02 & 0.229 \\
R3 & 33.25 & 0.520 \\
BART & 33.57 & 0.596   \\
\cmidrule{1-3}
MOVER & \textbf{39.43}  & \textbf{0.624} \\
\qquad w/o para score & 39.22  & 0.604  \\
\qquad w/o hypo ranker & 34.83  & 0.610 \\
\cmidrule{1-3}
COPY & \underline{51.69} & \underline{0.711} \\
\bottomrule
\end{tabular}
\caption{\label{auto-metric}
Automatic evaluation results on the test set of HYPO dataset.
}
\end{table}

\begin{table}
\setlength\tabcolsep{3pt}
\centering
\small
\begin{tabular}{lccccccccc}\toprule
\multirow{2}{*}{\makecell[c]{MOVER\\vs.}} &\multicolumn{2}{c}{Flu.} &\multicolumn{2}{c}{Hypo.} &\multicolumn{2}{c}{Crea.} &\multicolumn{2}{c}{Rel.} \\\cmidrule{2-9}
&W\% &L\% &W\% &L\% &W\% &L\% &W\% &L\% \\\cmidrule{1-9}
R1 &\textbf{79.7} &1.7 &\textbf{52.4} &47.6 &33.9 &\textbf{66.1} &\textbf{94.2} &4.3 \\
R3 &\textbf{35.8} &11.3 &\textbf{52.5} &36.1 &\textbf{50.0} &38.5 &\textbf{52.6} &29.8 \\
BART &\textbf{26.2} &19.7 &\textbf{67.7} &11.3 &\textbf{61.0} &10.2 &\textbf{49.2} &31.7 \\
HUMAN &\textbf{22.0} &18.6 &16.7 &\textbf{81.8} &14.3 &\textbf{84.3} &\textbf{46.8} &37.1 \\
\bottomrule
\end{tabular}
\caption{\label{human-metric}
Pairwise human comparison between MOVER and other baseline systems. Win[W]\% (Lose[L]\%) is the percentage of MOVER considered better (worse) than a baseline system. The rest are ties.
}
\end{table}

\subsection{Results}\label{sec:results}
\paragraph{Automatic Evaluation}
Table \ref{auto-metric} shows the automatic evaluation results of our system compared to different baselines. 
MOVER outperforms all three baselines on these two metrics.
However, BLEU and BERTScore are far not comprehensive evaluation measures for our hyperbole generation task, since there are only a few modifications from literal to hyperbole and thus there is a lot of overlap between the generated sentence and the source sentence. Even a naive system (COPY in Table \ref{auto-metric}) that simply returns the literal input verbatim as output \cite{krishna-etal-2020-reformulating} can achieve the highest performance. As a result, automatic metrics are not suitable for evaluating models that tend to copy input as output.

\begin{table}
\setlength\tabcolsep{3pt}
\small
\centering
\begin{tabular}{l>{\RaggedRight}p{4cm}cccc}\toprule
System &Sentence &F. &H. &C. &R. \\\cmidrule{1-6}
LITERAL &Being out of fashion is very bad. &- &- &- &- \\
MOVER &Being out of fashion is \textit{sheer hell}. &- &- &- &- \\
R1 & \textit{Their music will never go} out of fashion. &T &\textbf{W} &L &\textbf{W} \\
R3 &Being out of fashion is \textit{richly} bad. &T &\textbf{W} &\textbf{W} &T \\
BART &Being out of fashion is very \textit{difficult}. &T &\textbf{W} &\textbf{W} &T \\
HUMAN &\textit{Better be out of the world than} out of the fashion. &\textbf{W} &\textbf{W} &L &\textbf{W} \\

\bottomrule
\end{tabular}
\caption{Pairwise evaluation results (Win[W], Lose[L], Tie[T]) in terms of \textbf{F}luency, \textbf{H}yperbolicity, \textbf{C}reativity and \textbf{R}elevance between MOVER and the generated outputs of the baseline systems. Changed text spans are in \textit{italic}. More examples are given in Appendix \ref{appendix:more-examples}.}\label{tab:eval-example}
\end{table}

\begin{table*}
\centering
\begin{tabular}{lccc}
\toprule
Generated Hyperbole $s$ & $\text{hypo}(s)$ & $\text{para}(s)$ & $\text{score}(s)$ \\
\midrule
You have ravished me away by a power I \textit{cannot} resist. & 0.962 & 0.954 & 0.962 \\
You have ravished me away by a power I find \textit{unyielding} to resist.
 & 0.960  & 0.959 & 0.960 \\
You have ravished me \textit{alive} by a power I find difficult to resist. & 0.954 & 0.931 & 0.954 \\
You have \textit{driven} me away by a power I find difficult to resist. &  0.858	& 0.914  & 0.858 \\
You have ravished me away \textit{with a beauty} I find difficult to resist.  &  0.958 &	0.778 & 0.000 \\
\bottomrule
\end{tabular}%
\caption{\label{ranker}Intermediate results of the input literal sentence ``\textit{You have ravished me away by a power I find difficult to resist}'' after the over-generation steps (Section \ref{hypo-bart}). Their ranking scores (Section \ref{hypo-rank}) are displayed in the second to the fourth columns. Generated  hyperbolic text spans are in \textit{italic}.}
\end{table*}

\paragraph{Human Evaluation}
The inter-annotator agreement of raw human evaluation results in terms of Fleiss’ kappa \cite{fleiss1971measuring} is 0.212, which indicates fair agreement \cite{landis1977measurement}. We take a conservative approach and only consider items with an absolute majority label, i.e., at least three of the five labelers choose the same preference (win/lose/tie). On average there are 61 (86\%) items left for each baseline-criteria pair that satisfy this requirement. On this subset of items, Fleiss’ Kappa increases to 0.278 (fair agreement). This degree of agreement is acceptable compared to other sentence revision tasks (e.g., 0.322 by \citet{tan-lee-2014-corpus} and 0.263 by \citet{ afrin-litman-2018-annotation}) since it is hard to discern the subtle changing effect caused by local revision. 

The annotation results in Table \ref{human-metric} are the absolute majority vote (majority $>=$ 3) from the 5 annotators for each item. The results show that our model mostly outperforms (Win\% $>$ Lose\%) other baselines in the four metrics, except for creativity on R1. Because R1 directly retrieves human written hyperboles from HYPO-XL and is not strict about the relevance, it has the advantage of being more creative naturally. An example of this is shown in Table \ref{tab:eval-example}. Our model achieves a balance between generating hyperbolic output while preserving content, indicating the effectiveness of the ``overgenerate-and-rank'' mechanism. It is also worth noting that in terms of hyperbolicity, MOVER even performs better than human for 16.7\% of the test cases. Table \ref{tab:eval-example} shows a case where MOVER is rated higher than human.

\paragraph{Case Study}
Table \ref{tab:eval-example} shows a group of generated examples from different systems. MOVER changes the phrase ``very bad'' in the original input to an extreme expression ``sheer hell'', which captures the sentiment polarity of the original sentence while providing a hyperbolic effect. R1 retrieves a hyperbolic but irrelevant sentence. R3 replaces the word ``very'' with ``richly'', which is not coherent to the context, although the word ``richly'' may introduce some hyperbolic effects. BART just generates a literal sentence, which seems to be a simple paraphrase. Although human reference provides a valid hyperbolic paraphrase, the annotators prefer our version in terms of fluency, hyperbolicity and relevance. Since our system makes fewer edits to the input than to the human reference, we are more likely to win in fluency and relevance. Also, the generated hyperbolic span ``sheer hell'' presents a more extreme exaggeration than ``out of the world'' according to the human annotators. 

Table \ref{ranker} shows the over-generation results for a literal input, with their hyperbole and paraphrase scores. On the one hand,  our system can generate different hyperbolic versions, like the generated words ``cannot'', ``unyielding'', and ``alive''. This is reasonable since there might be multiple hyperbolic paraphrases for a single sentence. It is only for comparison with other baselines that we have to use the ranker to keep only one output, which inevitably undermines the strength of our approach. On the other hand, our ranker filters out the sentence if the infilled text violates the original meaning, which can be seen in the last row of Table \ref{ranker}. In this way, we gain explicit control over hyperbolicity and relevance through a scoring function, and endow MOVER with more explainability.

Despite the interesting results, we also observe the following types of errors in the generated outputs:
\begin{itemize}[itemsep=2pt,topsep=2pt,parsep=2pt]
\item The output is a paraphrase instead of hyperbole: ``\textit{My aim is very certain}'' $\rightarrow$ ``\textit{My aim is very \underline{clear}}''.
\item The degree of exaggeration is not enough: ``\textit{The news has been exaggerated}'' $\rightarrow$ ``\textit{The news has been \underline{greatly} exaggerated}''.
\item The output is not meaningful: ``\textit{I'd love to hang out every day}'' $\rightarrow$ ``\textit{I'd love to \underline{live} every day}''. We believe that incorporating more commonsense knowledge and generating freeform hyperboles beyond word- or phrase-level substitutions are promising for future improvement.
\end{itemize}

\paragraph{Ablation Study}
We investigate the impact of removing partial or all information during the ranking stage. The results are shown in Table \ref{auto-metric}. Specifically, if we rank multiple generated outputs by only hyperbole score (w/o para score), or randomly select one as the output (w/o hypo ranker), the performance will become worse. Note that we do not report the ablation result for ranking only by paraphrase score (w/o hypo score), because it has the same problem with COPY: a generated sentence that directly copies the input will result in the highest paraphrase score and thus be selected as the final output.

Furthermore, we note that the experiments on R3 and BART also serve as ablation studies for the text infilling model in Section \ref{hypo-bart} as they substitute the fine-tuned BART with a retrieve-and-replace method and a non-fine-tuned BART, respectively.

\section{Related Work}
\paragraph{Hyperbole Corpus}
\citet{troiano-etal-2018-computational} built the HYPO dataset consisting of 709 hyperbolic sentences with human-written paraphrases and lexically overlapping non-hyperbolic counterparts. \citet{kong-etal-2020-identifying} also built a Chinese hyperbole dataset with 2680 hyperboles. 
Our HYPO-L and HYPO-XL are substantially larger than HYPO and we hope they can facilitate computational research on hyperbole detection and generation.

\paragraph{Figurative Language Generation}
As a figure of speech, hyperbole generation is related to the general task of figurative language generation. Previous studies have tackled the generation of metaphor \cite{yu-wan-2019-avoid, stowe2020metaphoric, chakrabarty2021mermaid, stowe2021metaphor}, simile \cite{chakrabarty-etal-2020-generating,zhang2021writing},
idiom \cite{zhou2021solving},
pun \cite{yu2018neural,luo2019pun,he2019pun,yu2020homophonic},
and sarcasm \cite{chakrabarty-etal-2020-r}. HypoGen \cite{tian-etal-2021-hypogen-hyperbole} is a concurrent work with ours on hyperbole generation. However, we share a different point of view and the two methods are not directly comparable. They tackle the generation of \textit{clause-level} hyperboles and frame it as a sentence \textit{completion} task, while we focus on the \textit{word-level} or \textit{phrase-level} ones and frame it as a sentence \textit{editing} task. In addition, their collected hyperboles and generated outputs are limited to the  ``\textit{so...that}'' pattern while we do not posit constraints on sentence patterns.

\paragraph{Unsupervised Text Style Transfer}
Recent advances on unsupervised text style transfer \citep{hu2017toward, subramanian2018multiple, luo2019dual, zeng2020style} focus on transferring from one text attribute to another without parallel data. \citet{jin2020deep} classify existing methods into three main branches: \textit{disentanglement}, \textit{prototype editing}, and \textit{ pseudo-parallel corpus construction}. We argue that hyperbole generation is different from text style transfer. First, it is unclear whether ``literal'' and ``hyperbolic'' can be treated as ``styles'', especially the former one. Because ``literal'' sentences do not have any specific characteristics at all, there are no attribute markers \cite{li-etal-2018-delete} in the input sentences, and thus many text style transfer methods based on \textit{prototype editing} cannot work. 
Second, the hyperbolic span can be lexically separable from, yet strongly dependent on, the context (Section \ref{hypo-span}). On the contrary, \textit{disentanglement-based} approaches for text style transfer aim to separate content and style via latent representation learning. 
Third, MOVER could also be used for \textit{constructing pseudo-parallel corpus} of literal-hyperbole pairs given enough literal sentences as input, which is beyond the scope of this work.

\paragraph{Unsupervised Paraphrase Generation}
Unsupervised paraphrase generation models \cite{wieting2017learning,zhang2019syntax,roy2019unsupervised,huang2021generating} do not require paraphrase pairs for training. Although hyperbole generation also needs content preservation and lacks parallel training data, it is still different from paraphrase generation because we need to create a balance between paraphrasing and exaggerating. We further note that the task of metaphor generation \cite{chakrabarty2021mermaid}, which replaces a verb (e.g., ``\textit{The scream filled the night}'' $\rightarrow$ ``\textit{The scream \underline{pierced} the night}''), is also independent of paraphrase generation.

\section{Conclusion}
We tackle the challenging task of figurative language generation: hyperbole generation from literal sentences. We build the first large-scale hyperbole corpus HYPO-XL and propose an unsupervised approach MOVER for generating hyperbole in a controllable way. The results of automatic and human evaluation show that our model is successful in generating hyperbole. The proposed generation pipeline has better interpretability and flexibility compared to the potential end-to-end methods. In future, we plan to apply our ``mask-overgenerate-rank'' approach to the generation of other figurative languages, such as metaphor and irony.

\section{Ethical Consideration}

The HYPO-XL dataset is collected from a public website Sentencedict.com, and we have asked the website owners permission to use their data for research purposes. There is no explicit detail that leaks a user’s personal information including name, health, racial or ethnic origin, religious affiliation or beliefs, sexual orientation, etc.

Our proposed method MOVER utilizes the pretrained language model, which may inherit the bias in the massive training data. It is possible that MOVER is used for malicious purposes, since it does not explicitly filter input sentences with toxicity, bias or offensiveness. Therefore, the output generated by MOVER could potentially be harmful to certain groups or individuals. It is important that interested parties carefully address those biases before applying the model to real-world situations.

\section*{Acknowledgements}
This work was supported by National Science Foundation of China (No. 62161160339), National Key R\&D Program of China (No.2018YFB1005100), State Key Laboratory of Media Convergence Production Technology and Systems and Key Laboratory of Science, Techonology and Standard in Press Industry (Key Laboratory of Intelligent Press Media Technology). We appreciate the anonymous reviewers for their helpful comments. Xiaojun Wan is the corresponding auhor.

\bibliography{anthology,custom}

\begin{thebibliography}{47}
\expandafter\ifx\csname natexlab\endcsname\relax\def\natexlab#1{#1}\fi

\bibitem[{Afrin and Litman(2018)}]{afrin-litman-2018-annotation}
Tazin Afrin and Diane Litman. 2018.
\newblock \href {https://doi.org/10.18653/v1/W18-0528} {Annotation and
  classification of sentence-level revision improvement}.
\newblock In \emph{Proceedings of the Thirteenth Workshop on Innovative Use of
  {NLP} for Building Educational Applications}, pages 240--246, New Orleans,
  Louisiana. Association for Computational Linguistics.

\bibitem[{Chakrabarty et~al.(2020{\natexlab{a}})Chakrabarty, Ghosh, Muresan,
  and Peng}]{chakrabarty-etal-2020-r}
Tuhin Chakrabarty, Debanjan Ghosh, Smaranda Muresan, and Nanyun Peng.
  2020{\natexlab{a}}.
\newblock \href {https://doi.org/10.18653/v1/2020.acl-main.711} {{R}{\^{}}3:
  Reverse, retrieve, and rank for sarcasm generation with commonsense
  knowledge}.
\newblock In \emph{Proceedings of the 58th Annual Meeting of the Association
  for Computational Linguistics}, pages 7976--7986, Online. Association for
  Computational Linguistics.

\bibitem[{Chakrabarty et~al.(2020{\natexlab{b}})Chakrabarty, Muresan, and
  Peng}]{chakrabarty-etal-2020-generating}
Tuhin Chakrabarty, Smaranda Muresan, and Nanyun Peng. 2020{\natexlab{b}}.
\newblock \href {https://doi.org/10.18653/v1/2020.emnlp-main.524} {Generating
  similes effortlessly like a pro: A style transfer approach for simile
  generation}.
\newblock In \emph{Proceedings of the 2020 Conference on Empirical Methods in
  Natural Language Processing (EMNLP)}, pages 6455--6469, Online. Association
  for Computational Linguistics.

\bibitem[{Chakrabarty et~al.(2021)Chakrabarty, Zhang, Muresan, and
  Peng}]{chakrabarty2021mermaid}
Tuhin Chakrabarty, Xurui Zhang, Smaranda Muresan, and Nanyun Peng. 2021.
\newblock \href {https://doi.org/10.18653/v1/2021.naacl-main.336} {{MERMAID:}
  metaphor generation with symbolism and discriminative decoding}.
\newblock In \emph{Proceedings of the 2021 Conference of the North American
  Chapter of the Association for Computational Linguistics: Human Language
  Technologies, {NAACL-HLT} 2021, Online, June 6-11, 2021}, pages 4250--4261.
  Association for Computational Linguistics.

\bibitem[{Claridge(2010)}]{claridge2010hyperbole}
Claudia Claridge. 2010.
\newblock \emph{Hyperbole in English: A corpus-based study of exaggeration}.
\newblock Cambridge University Press.

\bibitem[{Devlin et~al.(2019)Devlin, Chang, Lee, and
  Toutanova}]{devlin-etal-2019-bert}
Jacob Devlin, Ming-Wei Chang, Kenton Lee, and Kristina Toutanova. 2019.
\newblock \href {https://doi.org/10.18653/v1/N19-1423} {{BERT}: Pre-training of
  deep bidirectional transformers for language understanding}.
\newblock In \emph{Proceedings of the 2019 Conference of the North {A}merican
  Chapter of the Association for Computational Linguistics: Human Language
  Technologies, Volume 1 (Long and Short Papers)}, pages 4171--4186,
  Minneapolis, Minnesota. Association for Computational Linguistics.

\bibitem[{Fleiss(1971)}]{fleiss1971measuring}
Joseph~L Fleiss. 1971.
\newblock Measuring nominal scale agreement among many raters.
\newblock \emph{Psychological bulletin}, 76(5):378.

\bibitem[{He et~al.(2019)He, Peng, and Liang}]{he2019pun}
He~He, Nanyun Peng, and Percy Liang. 2019.
\newblock Pun generation with surprise.
\newblock In \emph{Proceedings of the 2019 Conference of the North American
  Chapter of the Association for Computational Linguistics: Human Language
  Technologies, Volume 1 (Long and Short Papers)}, pages 1734--1744.

\bibitem[{Heilman and Smith(2009)}]{heilman2009question}
Michael Heilman and Noah~A Smith. 2009.
\newblock Question generation via overgenerating transformations and ranking.
\newblock Technical report, Carnegie-Mellon Univ Pittsburgh pa language
  technologies insT.

\bibitem[{Hu et~al.(2017)Hu, Yang, Liang, Salakhutdinov, and
  Xing}]{hu2017toward}
Zhiting Hu, Zichao Yang, Xiaodan Liang, Ruslan Salakhutdinov, and Eric~P Xing.
  2017.
\newblock Toward controlled generation of text.
\newblock In \emph{International Conference on Machine Learning}, pages
  1587--1596. PMLR.

\bibitem[{Huang and Chang(2021)}]{huang2021generating}
Kuan-Hao Huang and Kai-Wei Chang. 2021.
\newblock Generating syntactically controlled paraphrases without using
  annotated parallel pairs.
\newblock In \emph{Proceedings of the 16th Conference of the European Chapter
  of the Association for Computational Linguistics: Main Volume}, pages
  1022--1033.

\bibitem[{Jin et~al.(2020)Jin, Jin, Hu, Vechtomova, and Mihalcea}]{jin2020deep}
Di~Jin, Zhijing Jin, Zhiting Hu, Olga Vechtomova, and Rada Mihalcea. 2020.
\newblock Deep learning for text style transfer: A survey.
\newblock \emph{arXiv preprint arXiv:2011.00416}.

\bibitem[{Kong et~al.(2020)Kong, Li, Ge, Luo, and
  Ng}]{kong-etal-2020-identifying}
Li~Kong, Chuanyi Li, Jidong Ge, Bin Luo, and Vincent Ng. 2020.
\newblock \href {https://doi.org/10.18653/v1/2020.emnlp-main.571} {Identifying
  exaggerated language}.
\newblock In \emph{Proceedings of the 2020 Conference on Empirical Methods in
  Natural Language Processing (EMNLP)}, pages 7024--7034, Online. Association
  for Computational Linguistics.

\bibitem[{Kreuz and Roberts(1993)}]{KREUZ1993151}
Roger~J. Kreuz and Richard~M. Roberts. 1993.
\newblock \href {https://doi.org/https://doi.org/10.1016/0304-422X(93)90026-D}
  {The empirical study of figurative language in literature}.
\newblock \emph{Poetics}, 22(1):151--169.

\bibitem[{Krishna et~al.(2020)Krishna, Wieting, and
  Iyyer}]{krishna-etal-2020-reformulating}
Kalpesh Krishna, John Wieting, and Mohit Iyyer. 2020.
\newblock \href {https://doi.org/10.18653/v1/2020.emnlp-main.55} {Reformulating
  unsupervised style transfer as paraphrase generation}.
\newblock In \emph{Proceedings of the 2020 Conference on Empirical Methods in
  Natural Language Processing (EMNLP)}, pages 737--762, Online. Association for
  Computational Linguistics.

\bibitem[{Landis and Koch(1977)}]{landis1977measurement}
J~Richard Landis and Gary~G Koch. 1977.
\newblock The measurement of observer agreement for categorical data.
\newblock \emph{biometrics}, pages 159--174.

\bibitem[{Lewis et~al.(2020)Lewis, Liu, Goyal, Ghazvininejad, Mohamed, Levy,
  Stoyanov, and Zettlemoyer}]{lewis-etal-2020-bart}
Mike Lewis, Yinhan Liu, Naman Goyal, Marjan Ghazvininejad, Abdelrahman Mohamed,
  Omer Levy, Veselin Stoyanov, and Luke Zettlemoyer. 2020.
\newblock \href {https://doi.org/10.18653/v1/2020.acl-main.703} {{BART}:
  Denoising sequence-to-sequence pre-training for natural language generation,
  translation, and comprehension}.
\newblock In \emph{Proceedings of the 58th Annual Meeting of the Association
  for Computational Linguistics}, pages 7871--7880, Online. Association for
  Computational Linguistics.

\bibitem[{Li et~al.(2018)Li, Jia, He, and Liang}]{li-etal-2018-delete}
Juncen Li, Robin Jia, He~He, and Percy Liang. 2018.
\newblock \href {https://doi.org/10.18653/v1/N18-1169} {Delete, retrieve,
  generate: a simple approach to sentiment and style transfer}.
\newblock In \emph{Proceedings of the 2018 Conference of the North {A}merican
  Chapter of the Association for Computational Linguistics: Human Language
  Technologies, Volume 1 (Long Papers)}, pages 1865--1874, New Orleans,
  Louisiana. Association for Computational Linguistics.

\bibitem[{Luo et~al.(2019{\natexlab{a}})Luo, Li, Zhou, Yang, Chang, Sun, and
  Sui}]{luo2019dual}
Fuli Luo, Peng Li, Jie Zhou, Pengcheng Yang, Baobao Chang, Xu~Sun, and Zhifang
  Sui. 2019{\natexlab{a}}.
\newblock \href {https://doi.org/10.24963/ijcai.2019/711} {A dual reinforcement
  learning framework for unsupervised text style transfer}.
\newblock In \emph{Proceedings of the Twenty-Eighth International Joint
  Conference on Artificial Intelligence, {IJCAI} 2019, Macao, China, August
  10-16, 2019}, pages 5116--5122. ijcai.org.

\bibitem[{Luo et~al.(2019{\natexlab{b}})Luo, Li, Yang, Li, Chang, Sui, and
  Sun}]{luo2019pun}
Fuli Luo, Shunyao Li, Pengcheng Yang, Lei Li, Baobao Chang, Zhifang Sui, and
  Xu~Sun. 2019{\natexlab{b}}.
\newblock Pun-gan: Generative adversarial network for pun generation.
\newblock In \emph{Proceedings of the 2019 Conference on Empirical Methods in
  Natural Language Processing and the 9th International Joint Conference on
  Natural Language Processing (EMNLP-IJCNLP)}, pages 3388--3393.

\bibitem[{Manning et~al.(2014)Manning, Surdeanu, Bauer, Finkel, Bethard, and
  McClosky}]{manning-etal-2014-stanford}
Christopher Manning, Mihai Surdeanu, John Bauer, Jenny Finkel, Steven Bethard,
  and David McClosky. 2014.
\newblock \href {https://doi.org/10.3115/v1/P14-5010} {The {S}tanford
  {C}ore{NLP} natural language processing toolkit}.
\newblock In \emph{Proceedings of 52nd Annual Meeting of the Association for
  Computational Linguistics: System Demonstrations}, pages 55--60, Baltimore,
  Maryland. Association for Computational Linguistics.

\bibitem[{Nikolov et~al.(2020)Nikolov, Malmi, Northcutt, and
  Parisi}]{nikolov2020rapformer}
Nikola~I Nikolov, Eric Malmi, Curtis Northcutt, and Loreto Parisi. 2020.
\newblock Rapformer: Conditional rap lyrics generation with denoising
  autoencoders.
\newblock In \emph{Proceedings of the 13th International Conference on Natural
  Language Generation}, pages 360--373.

\bibitem[{Novikova et~al.(2017)Novikova, Du{\v{s}}ek, Cercas~Curry, and
  Rieser}]{novikova-etal-2017-need}
Jekaterina Novikova, Ond{\v{r}}ej Du{\v{s}}ek, Amanda Cercas~Curry, and Verena
  Rieser. 2017.
\newblock \href {https://doi.org/10.18653/v1/D17-1238} {Why we need new
  evaluation metrics for {NLG}}.
\newblock In \emph{Proceedings of the 2017 Conference on Empirical Methods in
  Natural Language Processing}, pages 2241--2252, Copenhagen, Denmark.
  Association for Computational Linguistics.

\bibitem[{Papineni et~al.(2002)Papineni, Roukos, Ward, and
  Zhu}]{papineni2002bleu}
Kishore Papineni, Salim Roukos, Todd Ward, and Wei-Jing Zhu. 2002.
\newblock Bleu: a method for automatic evaluation of machine translation.
\newblock In \emph{Proceedings of the 40th annual meeting of the Association
  for Computational Linguistics}, pages 311--318.

\bibitem[{Pennington et~al.(2014)Pennington, Socher, and
  Manning}]{pennington-etal-2014-glove}
Jeffrey Pennington, Richard Socher, and Christopher Manning. 2014.
\newblock \href {https://doi.org/10.3115/v1/D14-1162} {{G}lo{V}e: Global
  vectors for word representation}.
\newblock In \emph{Proceedings of the 2014 Conference on Empirical Methods in
  Natural Language Processing ({EMNLP})}, pages 1532--1543, Doha, Qatar.
  Association for Computational Linguistics.

\bibitem[{Rajpurkar et~al.(2016)Rajpurkar, Zhang, Lopyrev, and
  Liang}]{rajpurkar-etal-2016-squad}
Pranav Rajpurkar, Jian Zhang, Konstantin Lopyrev, and Percy Liang. 2016.
\newblock \href {https://doi.org/10.18653/v1/D16-1264} {{SQ}u{AD}: 100,000+
  questions for machine comprehension of text}.
\newblock In \emph{Proceedings of the 2016 Conference on Empirical Methods in
  Natural Language Processing}, pages 2383--2392, Austin, Texas. Association
  for Computational Linguistics.

\bibitem[{Reimers and Gurevych(2019)}]{reimers-2019-sentence-bert}
Nils Reimers and Iryna Gurevych. 2019.
\newblock \href {https://arxiv.org/abs/1908.10084} {Sentence-bert: Sentence
  embeddings using siamese bert-networks}.
\newblock In \emph{Proceedings of the 2019 Conference on Empirical Methods in
  Natural Language Processing}. Association for Computational Linguistics.

\bibitem[{Roy and Grangier(2019)}]{roy2019unsupervised}
Aurko Roy and David Grangier. 2019.
\newblock Unsupervised paraphrasing without translation.
\newblock In \emph{Proceedings of the 57th Annual Meeting of the Association
  for Computational Linguistics}, pages 6033--6039.

\bibitem[{Shao et~al.(2019)Shao, Huang, Wen, Xu, and Zhu}]{shao-etal-2019-long}
Zhihong Shao, Minlie Huang, Jiangtao Wen, Wenfei Xu, and Xiaoyan Zhu. 2019.
\newblock \href {https://doi.org/10.18653/v1/D19-1321} {Long and diverse text
  generation with planning-based hierarchical variational model}.
\newblock In \emph{Proceedings of the 2019 Conference on Empirical Methods in
  Natural Language Processing and the 9th International Joint Conference on
  Natural Language Processing (EMNLP-IJCNLP)}, pages 3257--3268, Hong Kong,
  China. Association for Computational Linguistics.

\bibitem[{Stowe et~al.(2021)Stowe, Chakrabarty, Peng, Muresan, and
  Gurevych}]{stowe2021metaphor}
Kevin Stowe, Tuhin Chakrabarty, Nanyun Peng, Smaranda Muresan, and Iryna
  Gurevych. 2021.
\newblock \href {https://doi.org/10.18653/v1/2021.acl-long.524} {Metaphor
  generation with conceptual mappings}.
\newblock In \emph{Proceedings of the 59th Annual Meeting of the Association
  for Computational Linguistics and the 11th International Joint Conference on
  Natural Language Processing, {ACL/IJCNLP} 2021, (Volume 1: Long Papers),
  Virtual Event, August 1-6, 2021}, pages 6724--6736. Association for
  Computational Linguistics.

\bibitem[{Stowe et~al.(2020)Stowe, Ribeiro, and Gurevych}]{stowe2020metaphoric}
Kevin Stowe, Leonardo Ribeiro, and Iryna Gurevych. 2020.
\newblock Metaphoric paraphrase generation.
\newblock \emph{arXiv preprint arXiv:2002.12854}.

\bibitem[{Subramanian et~al.(2018)Subramanian, Lample, Smith, Denoyer, Ranzato,
  and Boureau}]{subramanian2018multiple}
Sandeep Subramanian, Guillaume Lample, Eric~Michael Smith, Ludovic Denoyer,
  Marc'Aurelio Ranzato, and Y-Lan Boureau. 2018.
\newblock Multiple-attribute text style transfer.
\newblock \emph{arXiv preprint arXiv:1811.00552}.

\bibitem[{Tan and Lee(2014)}]{tan-lee-2014-corpus}
Chenhao Tan and Lillian Lee. 2014.
\newblock \href {https://doi.org/10.3115/v1/P14-2066} {A corpus of
  sentence-level revisions in academic writing: A step towards understanding
  statement strength in communication}.
\newblock In \emph{Proceedings of the 52nd Annual Meeting of the Association
  for Computational Linguistics (Volume 2: Short Papers)}, pages 403--408,
  Baltimore, Maryland. Association for Computational Linguistics.

\bibitem[{Tian et~al.(2021)Tian, Sridhar, and
  Peng}]{tian-etal-2021-hypogen-hyperbole}
Yufei Tian, Arvind~krishna Sridhar, and Nanyun Peng. 2021.
\newblock \href {https://aclanthology.org/2021.findings-emnlp.136}
  {{H}ypo{G}en: Hyperbole generation with commonsense and counterfactual
  knowledge}.
\newblock In \emph{Findings of the Association for Computational Linguistics:
  EMNLP 2021}, pages 1583--1593, Punta Cana, Dominican Republic. Association
  for Computational Linguistics.

\bibitem[{Troiano et~al.(2018)Troiano, Strapparava, {\"O}zbal, and
  Tekiro{\u{g}}lu}]{troiano-etal-2018-computational}
Enrica Troiano, Carlo Strapparava, G{\"o}zde {\"O}zbal, and Serra~Sinem
  Tekiro{\u{g}}lu. 2018.
\newblock \href {https://doi.org/10.18653/v1/D18-1367} {A computational
  exploration of exaggeration}.
\newblock In \emph{Proceedings of the 2018 Conference on Empirical Methods in
  Natural Language Processing}, pages 3296--3304, Brussels, Belgium.
  Association for Computational Linguistics.

\bibitem[{Wang et~al.(2018)Wang, Singh, Michael, Hill, Levy, and
  Bowman}]{wang-etal-2018-glue}
Alex Wang, Amanpreet Singh, Julian Michael, Felix Hill, Omer Levy, and Samuel
  Bowman. 2018.
\newblock \href {https://doi.org/10.18653/v1/W18-5446} {{GLUE}: A multi-task
  benchmark and analysis platform for natural language understanding}.
\newblock In \emph{Proceedings of the 2018 {EMNLP} Workshop {B}lackbox{NLP}:
  Analyzing and Interpreting Neural Networks for {NLP}}, pages 353--355,
  Brussels, Belgium. Association for Computational Linguistics.

\bibitem[{Wieting et~al.(2017)Wieting, Mallinson, and
  Gimpel}]{wieting2017learning}
John Wieting, Jonathan Mallinson, and Kevin Gimpel. 2017.
\newblock Learning paraphrastic sentence embeddings from back-translated
  bitext.
\newblock In \emph{Proceedings of the 2017 Conference on Empirical Methods in
  Natural Language Processing}, pages 274--285.

\bibitem[{Wolf et~al.(2020)Wolf, Debut, Sanh, Chaumond, Delangue, Moi, Cistac,
  Rault, Louf, Funtowicz, Davison, Shleifer, von Platen, Ma, Jernite, Plu, Xu,
  Le~Scao, Gugger, Drame, Lhoest, and Rush}]{wolf-etal-2020-transformers}
Thomas Wolf, Lysandre Debut, Victor Sanh, Julien Chaumond, Clement Delangue,
  Anthony Moi, Pierric Cistac, Tim Rault, Remi Louf, Morgan Funtowicz, Joe
  Davison, Sam Shleifer, Patrick von Platen, Clara Ma, Yacine Jernite, Julien
  Plu, Canwen Xu, Teven Le~Scao, Sylvain Gugger, Mariama Drame, Quentin Lhoest,
  and Alexander Rush. 2020.
\newblock \href {https://doi.org/10.18653/v1/2020.emnlp-demos.6} {Transformers:
  State-of-the-art natural language processing}.
\newblock In \emph{Proceedings of the 2020 Conference on Empirical Methods in
  Natural Language Processing: System Demonstrations}, pages 38--45, Online.
  Association for Computational Linguistics.

\bibitem[{Yu et~al.(2018)Yu, Tan, and Wan}]{yu2018neural}
Zhiwei Yu, Jiwei Tan, and Xiaojun Wan. 2018.
\newblock A neural approach to pun generation.
\newblock In \emph{Proceedings of the 56th Annual Meeting of the Association
  for Computational Linguistics (Volume 1: Long Papers)}, pages 1650--1660.

\bibitem[{Yu and Wan(2019)}]{yu-wan-2019-avoid}
Zhiwei Yu and Xiaojun Wan. 2019.
\newblock \href {https://doi.org/10.18653/v1/N19-1092} {How to avoid sentences
  spelling boring? towards a neural approach to unsupervised metaphor
  generation}.
\newblock In \emph{Proceedings of the 2019 Conference of the North {A}merican
  Chapter of the Association for Computational Linguistics: Human Language
  Technologies, Volume 1 (Long and Short Papers)}, pages 861--871, Minneapolis,
  Minnesota. Association for Computational Linguistics.

\bibitem[{Yu et~al.(2020)Yu, Zang, and Wan}]{yu2020homophonic}
Zhiwei Yu, Hongyu Zang, and Xiaojun Wan. 2020.
\newblock Homophonic pun generation with lexically constrained rewriting.
\newblock In \emph{Proceedings of the 2020 Conference on Empirical Methods in
  Natural Language Processing (EMNLP)}, pages 2870--2876.

\bibitem[{Zeng et~al.(2020)Zeng, Shoeybi, and Liu}]{zeng2020style}
Kuo-Hao Zeng, Mohammad Shoeybi, and Ming-Yu Liu. 2020.
\newblock Style example-guided text generation using generative adversarial
  transformers.
\newblock \emph{arXiv preprint arXiv:2003.00674}.

\bibitem[{Zhang et~al.(2021)Zhang, Cui, Xia, Guo, Li, Wei, and
  Cui}]{zhang2021writing}
Jiayi Zhang, Zhi Cui, Xiaoqiang Xia, Yalong Guo, Yanran Li, Chen Wei, and
  Jianwei Cui. 2021.
\newblock Writing polishment with simile: Task, dataset and a neural approach.
\newblock In \emph{Proceedings of the AAAI Conference on Artificial
  Intelligence}, volume~35, pages 14383--14392.

\bibitem[{Zhang et~al.(2020)Zhang, Kishore, Wu, Weinberger, and
  Artzi}]{zhang2019bertscore}
Tianyi Zhang, Varsha Kishore, Felix Wu, Kilian~Q. Weinberger, and Yoav Artzi.
  2020.
\newblock \href {https://openreview.net/forum?id=SkeHuCVFDr} {Bertscore:
  Evaluating text generation with {BERT}}.
\newblock In \emph{8th International Conference on Learning Representations,
  {ICLR} 2020, Addis Ababa, Ethiopia, April 26-30, 2020}. OpenReview.net.

\bibitem[{Zhang et~al.(2019{\natexlab{a}})Zhang, Yang, Yuan, Shen, and
  Carin}]{zhang2019syntax}
Xinyuan Zhang, Yi~Yang, Siyang Yuan, Dinghan Shen, and Lawrence Carin.
  2019{\natexlab{a}}.
\newblock Syntax-infused variational autoencoder for text generation.
\newblock In \emph{Proceedings of the 57th Annual Meeting of the Association
  for Computational Linguistics}, pages 2069--2078.

\bibitem[{Zhang et~al.(2019{\natexlab{b}})Zhang, Baldridge, and
  He}]{zhang2019paws}
Yuan Zhang, Jason Baldridge, and Luheng He. 2019{\natexlab{b}}.
\newblock \href {https://doi.org/10.18653/v1/n19-1131} {{PAWS:} paraphrase
  adversaries from word scrambling}.
\newblock In \emph{Proceedings of the 2019 Conference of the North American
  Chapter of the Association for Computational Linguistics: Human Language
  Technologies, {NAACL-HLT} 2019, Minneapolis, MN, USA, June 2-7, 2019, Volume
  1 (Long and Short Papers)}, pages 1298--1308. Association for Computational
  Linguistics.

\bibitem[{Zhou et~al.(2021)Zhou, Gong, Nanniyur, and Bhat}]{zhou2021solving}
Jianing Zhou, Hongyu Gong, Srihari Nanniyur, and Suma Bhat. 2021.
\newblock From solving a problem boldly to cutting the gordian knot: Idiomatic
  text generation.
\newblock \emph{arXiv preprint arXiv:2104.06541}.

\end{thebibliography}
\bibliographystyle{acl_natbib}
\appendix

\section{Additional Dataset Statistics}\label{sec:dataset-stat}

We annotate the hyperbolic spans (Section \ref{hypo-span}) for the 94 real hyperboles in Section \ref{sec:corpus-analysis} and show some examples of the most common POS n-grams of hyperbolic spans in Table \ref{tab:pos-example}. We further follow \citet{troiano-etal-2018-computational} to annotate the types of exaggeration along three dimensions: ``measurable'', ``possible'' and ``conventional''.
A hyperbole is ``measurable'' if it exaggerates something which is objective and quantifiable. A hyperbole is rated as ``possible'' if it denotes an extreme but conceivable situation. A hyperbole is judged as ``conventional'' if it does not express an idea in a creative way. However, we note that there are no absolute answers for these three questions and the annotation results may be subjective. Each hyperbole is either YES or NO for each dimension and the reported numbers in Table \ref{tab:type-example} are for YES. 

\section{More Generated Examples}\label{appendix:more-examples}
Table \ref{tab:more-eval-result} shows more examples of output generated from different systems and human references.

\begin{table*}
\centering
\begin{tabular}{cc>{\RaggedRight}p{8.5cm}}\toprule
POS &\#  & Hyperbole Example \\\cmidrule{1-3}
NN &19  &His words confirmed \textit{everything}. \\
RB &15  &He descanted \textit{endlessly} upon the wonders of his trip. \\
JJ &14  &Youth means \textit{limitless} possibilities. \\
\bottomrule
\end{tabular}
\caption{Three most common POS n-grams of hyperbolic spans in 94 randomly sampled hyperboles from HYPO-XL. The hyperbolic spans are in \textit{italic}.}\label{tab:pos-example}
\end{table*}

\begin{table*}
\centering
\begin{tabular}{cc>{\RaggedRight}p{10cm}}\toprule
Type &\#  &Hyperbole Example \\\cmidrule{1-3}
Measurable & 44 & At any moment, I feared, the boys could \textit{snap my body in half} with just one concerted shove. \\
Possible & 27 & The words caused a shiver to \textit{run a fine sharp line} through her. \\
Conventional & 65 & She is \textit{forever} picking at the child. \\
\bottomrule
\end{tabular}
\caption{Three types of exaggeration in 94 randomly sampled hyperboles from HYPO-XL. The hyperbolic spans are in \textit{italic}.}\label{tab:type-example}
\end{table*}




\begin{table*}
\setlength\tabcolsep{3pt}
\centering
\begin{tabular}{l>{\RaggedRight}p{10.2cm}cccc}\toprule
System &Sentence &Flu. &Hypo. &Crea. &Rel. \\\cmidrule{1-6}
LITERAL &At that point, the presidency was hard to recover. &- &- &- &- \\
MOVER &At that point the presidency was \textit{virtually impossible} to recover.  &- &- &- &-  \\
R1 &\textit{The destruction of a President with its collapse of executive authority was too staggering to contemplate}. &\textbf{W} &\textbf{W} &T &\textbf{W} \\
R3 &At that point the presidency was \textit{staggering} to recover. &\textbf{W} &\textbf{W} &L &\textbf{W} \\
BART &At that point the presidency was \textit{too fragile} to recover &T &\textbf{W} &T &T \\
HUMAN &At that point, the presidency was \textit{fatally wounded}. &T &\textbf{W} &\textbf{W} &\textbf{W} \\\cmidrule{1-6}
LITERAL &His piano playing is very bad. &- &- &- &- \\
MOVER &His piano playing is \textit{beyond} bad. &- &- &- &- \\
R1 &Her piano playing is \textit{absolute magic}. &T &T &L &\textbf{W} \\
R3 &His piano \textit{guitar} is very bad. &T &T &L &L \\
BART &His piano playing is very \textit{good}. &T &\textbf{W} &\textbf{W} &\textbf{W} \\
HUMAN &His piano playing is \textit{enough to make Beethoven turn in his grave}. &T &L &L &\textbf{W} \\\midrule
LITERAL &The professor humiliated me in front of the class. &- &- &- &- \\
MOVER &The professor humiliated me in \textit{every conceivable way}. &- &- &- &- \\
R1 &\textit{She infected the whole class with her enthusiasm.} &\textbf{W} &\textbf{W} &\textbf{W} &\textbf{W} \\
R3 &\textit{That lecture} humiliated me in front of the class. &T &\textbf{W} &T &T \\
BART &The professor humiliated me \textit{and the rest} of the class. &\textbf{W} &\textbf{W} &\textbf{W} &\textbf{W} \\
HUMAN &The professor \textit{destroyed} me in front of the class. &T &L &\textbf{W} &\textbf{W} \\\midrule
LITERAL &It annoys me when you only drink half of the soda. &- &- &- &- \\
MOVER &It \textit{kills} me when you only drink half of the soda. &- &- &- &- \\
R1 &\textit{That was the best ice-cream soda I ever tasted}. &T &\textbf{W} &\textbf{W} &\textbf{W} \\
R3 &It annoys me when you only drink \textit{boredom} of the soda. &T &\textbf{W} &\textbf{W} &T \\
BART &It annoys me when you only drink half of \textit{it}. &\textbf{W} &\textbf{W} &\textbf{W} &\textbf{W} \\
HUMAN &It \textit{drives me crazy} when you only drink half of the soda. &T &\textbf{W} &T &T \\
\bottomrule
\end{tabular}
\caption{Results of the pairwise evaluation (Win [W], Lose [L], Tie [T]) between MOVER and the outputs generated by the baseline systems. The changed text spans are in \textit{italic}.}\label{tab:more-eval-result}
\end{table*}

\end{document}